\documentclass[runningheads]{llncs}

\usepackage{algorithm}
\usepackage[noend]{algpseudocode}
\usepackage{amsmath}
\usepackage{graphicx}
\usepackage{float}
\usepackage{textcomp}
\usepackage{booktabs}
\usepackage[table,xcdraw]{xcolor}
\usepackage{adjustbox}

\begin{document}
\title{An Integrated System for Spatio-Temporal Summarization of 360-degrees Videos\thanks{This work was supported by the EU Horizon Europe and Horizon 2020 programmes under grant agreements 101070109 TransMIXR and 951911 AI4Media, respectively. Code is publicly-available at: https://github.com/IDT-ITI/CA-SUM-360.}}
\author{Ioannis Kontostathis\inst{}\orcidID{0009-0007-5311-8167} \and
Evlampios Apostolidis\inst{}\orcidID{0000-0001-5376-7158} \and
Vasileios Mezaris\inst{}\orcidID{0000-0002-0121-4364}}
\authorrunning{I. Kontostathis et al.}
\institute{Information Technologies Institute - Centre for Research and Technologies, Hellas, 6th km Charilaou - Thermi Road, 57001, Thessaloniki, Greece\\
\email{\{ioankont,apostolid,bmezaris\}@iti.gr}}
\maketitle
\begin{abstract}
In this work, we present an integrated system for spatiotemporal summarization of 360-degrees videos. The video summary production mainly involves the detection of salient events and their synopsis into a concise summary. The analysis relies on state-of-the-art methods for saliency detection in 360-degrees video (ATSal and SST-Sal) and video summarization (CA-SUM). It also contains a mechanism that classifies a 360-degrees video based on the use of static or moving camera during recording and decides which saliency detection method will be used, as well as a 2D video production component that is responsible to create a conventional 2D video containing the salient events in the 360-degrees video. Quantitative evaluations using two datasets for 360-degrees video saliency detection (VR-EyeTracking, Sports-360) show the accuracy and positive impact of the developed decision mechanism, and justify our choice to use two different methods for detecting the salient events. A qualitative analysis using content from these datasets, gives further insights about the functionality of the decision mechanism, shows the pros and cons of each used saliency detection method and demonstrates the advanced performance of the trained summarization method against a more conventional approach.

\end{abstract}
\section{Introduction}

Over the last years, we are experiencing a rapid growth of $360^{\circ}$ videos. This growth is mainly fueled by the increasing engagement of users with advanced $360^{\circ}$ video recording devices, such as GoPro and GearVR, the ability to share such content via video sharing platforms (such as YouTube and Vimeo) and social networks (such as Facebook), as well as the emergence of extended reality technologies. The rapid growth of $360^{\circ}$ video content comes with an increasing need for technologies that would allow the creation of a short summary that conveys information about the salient events of the video. Such technologies would allow users to quickly get a grasp about the content of the $360^{\circ}$ video, thus significantly facilitating their browsing and navigation in large collections.  

Despite the increasing popularity of $360^{\circ}$ video content, the research on the summarization of this content is still limited. A few methods focus on piloting the viewer through the unlimited field of view of the $360^{\circ}$ video, by controlling the position and the field of view of the camera and generating an optimal camera trajectory \cite{su2016activity,Su_2017_CVPR,Hu_2017_CVPR,9072511,10.1145/3306346.3323046}. These methods perform a spatial summarization on the $360^{\circ}$ video by focusing on the most salient regions and ignoring areas of the $360^{\circ}$ video that are less interesting. Nevertheless, the produced normal Field-Of-View (NFOV) video might contain redundant information as it presents the detected salient events in their full duration; thus it cannot be seen as a condensed summary of the video content. A couple of recent methods target both spatial and temporal summarization of $360^{\circ}$ videos \cite{Yu2019DeepRanking,Lee_2018_CVPR}. However, both methods assume the existence of a single important event \cite{Yu2019DeepRanking} or narrative \cite{Lee_2018_CVPR} in order to create a video highlight and a story-based video summary, respectively. 

In this paper, we propose an approach that considers both the spatial and the temporal dimension of the $360^{\circ}$ video (contrary to \cite{su2016activity,Su_2017_CVPR,Hu_2017_CVPR,9072511}) and takes into account several different events that might take place in parallel (differently to \cite{Yu2019DeepRanking,Lee_2018_CVPR}), in order to form a summary of the video content. The developed system is compatible with $360^{\circ}$ videos captured using either static or moving camera. Moreover, the duration of the output video summary can be adjusted according to the users' needs, thus facilitating the production of different summaries for a given $360^{\circ}$ video. Our main contributions are as follows:
\begin{itemize}
    \item We propose a new approach for spatiotemporal summarization of $360^{\circ}$ videos, that relies on a combination of state-of-the-art deep learning methods for saliency detection and video summarization.
    \item We develop a mechanism that classifies a $360^{\circ}$ video according to the use of a static or moving camera, and a method that forms a conventional 2D video showing the detected salient events in the $360^{\circ}$ video.
    \item We build an integrated system that performs spatiotemporal summarization of $360^{\circ}$ video in an end-to-end manner.
\end{itemize}

\section{Related Work}

One of the early attempts to offer a more natural-looking NFOV video that focuses on the interesting areas/events of a panoramic $360^{\circ}$ video, was made by Su et al. (2016) \cite{su2016activity}. Their method, called AutoCam, learns a discriminative model of human-captured NFOV Web videos and utilizes this model to identify candidate view-points and events of interest in a $360^{\circ}$ video. Then, it uses dynamic programming to stitch them together through optimal human-like camera motions and create a new NFOV presentation of the $360^{\circ}$ video content. In their following work, Su et al. (2017) \cite{Su_2017_CVPR}, generalized the problem by allowing the method to control the field of view of the camera dynamically (instead of keeping it fixed), applied a coarse-to-fine optimization approach that iteratively refines and makes it tractable, and integrated a mechanism to encourage diversity to the NFOV videos. Hu et al. (2017) \cite{Hu_2017_CVPR}, formulated the selection of the most interesting parts of a $360^{\circ}$ video as a $360^{\circ}$ viewer piloting task and described a deep-learning-based agent which uses an object detector \cite{NIPS2015_14bfa6bb} to extract a set of candidate objects of interest from each frame, and a trainable RNN to select the main object. Given the selected main object and the previously selected viewing angles, the agent shifts the current viewing angle to the next preferred one. The training process rewards the selection of the correct main object, aims to minimize the distance between the selected and ground-truth viewing angles, and tries to maximize the smoothness when transitioning between different viewing angles. On a similar direction, Qiao et al. (2017) \cite{9072511}, presented the multi-task DNN (MT-DNN) method that predicts viewport-dependent saliency over $360^{\circ}$ videos based on both video content and viewport location (the fraction of a $360^{\circ}$ scene that can be viewed by an observer). Each different task is associated with a different viewport, and for each task MT-DNN uses a combination of CNN and ConvLSTM for modeling both spatial and temporal features at specific viewport locations. The output of all tasks is fused by the overall MT-DNN that makes estimates about the saliency at any viewport location. Yu et al. (2018) \cite{Yu2019DeepRanking}, addressed the problem of $360^{\circ}$ video highlight detection. After defining NFOV segments, their method uses a trainable deep ranking model which produces a spherical score map of composition per video segment and determines which view can be considered as a highlight via a sliding window kernel. Based on the composition score map, their method performs spatial summarization by finding out the best NFOV subshot per $360^{\circ}$ video segment, and temporal summarization by selecting the N top-ranked NFOV subshots as a highlight for the entire $360^{\circ}$ video. A similar approach was proposed by Lee et al. (2018) \cite{Lee_2018_CVPR}, for story-based summarization of $360^{\circ}$ videos. It uses a trainable deep ranking network to score NFOV region proposals cropped from the input $360^{\circ}$ video. Then, it performs temporal summarization using a memory network that models the correlation between past and future information (video subshots), based on the assumption that the parts of a story-based summary share a common story-line. Finally, Kang et al. (2019) \cite{10.1145/3306346.3323046} presented an interactive $360^{\circ}$ video navigation system which applies optical flow and saliency detection to find a virtual camera path with the most salient events in the video, and generate a NFOV video.

\section{Proposed Approach}

\begin{figure}[t]
    \centering
    \includegraphics[width=\textwidth]{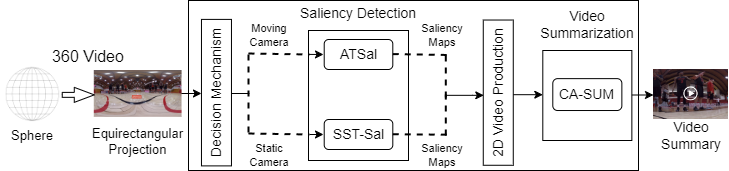}
    \caption{An overview of the proposed approach for $360^{\circ}$ video summarization. Dashed lines indicate alternative paths of the processing pipeline.}
    \label{fig:schmema}
\end{figure}

An overview of the proposed approach is given in Fig. \ref{fig:schmema}. The input $360^{\circ}$ video is initially subjected to equirectangular projection (ERP) to form a set of omnidirectional planar frames (called ERP frames in the following). This set of frames is then analysed by a mechanism that makes a decision on whether the $360^{\circ}$ video has been captured by a static or a moving camera. The use of such a decision mechanism in combination with two different methods for saliency detection was based on our study of the relevant literature and the observation that most methods either deal with or are more effective on $360^{\circ}$ videos that have been captured by one of the aforementioned video recording conditions. So, based on the output of the decision mechanism the ERP frames are subsequently forwarded to one of the integrated methods for saliency detection, which produce a set of frame-level saliency maps. The ERP frames and the extracted saliency maps are then given as input to a component that forms a 2D video containing the detected salient events in the $360^{\circ}$ video. Finally, the produced 2D video is processed by a video summarization method which makes estimates about the importance of each frame of the video and forms the video summary. More details about each processing component are given in the following subsections.

\subsection{Decision Mechanism}

\begin{figure}[t]
    \centering
    \includegraphics[width=0.6\textwidth]{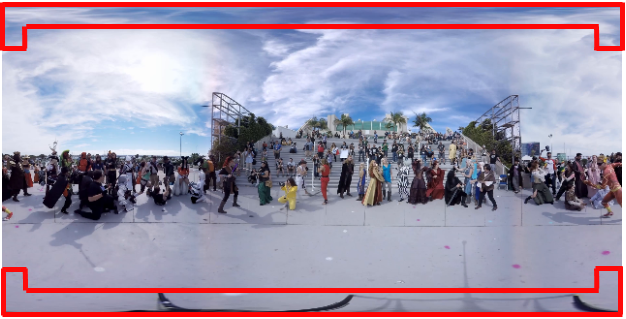}
    \caption{An ERP frame and its north and south regions (highlighted by the red-coloured bounding boxes) that are used by the developed decision mechanism.}
    \label{fig:decision_mech}
\end{figure}

As noted in \cite{Equator_2018_SPIC}, the main action zones on ERP frames are most commonly near the equator. Taking this into account, the decision on whether the $360^{\circ}$ video was captured by a static or moving camera is based on the analysis of the north and south regions of the ERP frames, as depicted in Fig. \ref{fig:decision_mech}. Such regions should exhibit limited variation across a sequence of frames when the $360^{\circ}$ video is captured by a static camera. So, given the ERP frames of the input video, the decision mechanism focuses on the aforementioned frame regions and computes the phase correlation \cite{Phasecorr_2002}. If the computed scores frequently exceed an experimentally defined threshold $t_0$, the mechanism declares the use of a moving camera; otherwise, it indicates the use of a static one.

\subsection{Saliency Detection}
\label{subsec:saliency_prediction}

Based on a literature review and experimentation with a few approaches, we use two state-of-the-art methods for saliency detection in $360^{\circ}$ videos, namely the ATSal \cite{Atsal_2020} and the SST-Sal method \cite{Sstsal_2022_CG}. ATSal \cite{Atsal_2020} employs two parallel models: i) an attention model that encodes global visual features from the ERP frame, and ii) the SalEMA expert model that focuses on temporal characteristics and uses cubemap projection (CMP) frames. The attention model consists of a fine-tuned VGG16 \cite{Vgg16_2015_model} encoder-decoder with an intermediate attention mechanism. The SalEMA expert model extracts the temporal characteristics of the $360^{\circ}$ video via CMP frames. It is composed of the SalEMA-Poles model, that is responsible for the north and south regions of the $360^{\circ}$ video, and the SalEMA-Equator model that focused on the front, back, left and right regions of the $360^{\circ}$ video. After feeding the ERP frame to the attention model and the CMP frame to the SalEMA expert model, the final saliency map is formulated after performing a pixel-wise multiplication between the outputs. SST-Sal \cite{Sstsal_2022_CG} adopts an encoder–decoder approach that takes into account both temporal and spatial information at feature encoding and decoding time. The encoder is formed by a Spherical Convolutional LSTM (ConvLSTM) and a spherical max pooling layer, and extracts the spatio-temporal features from an input sequence of ERP frames. The decoder is composed of a Spherical ConvLSTM and an up-sampling layer, and leverages the latent information to predict a sequence of saliency maps. The utilized ConvLSTMs extend the functionality of LSTMs by applying spherical convolutions, thus allowing to account for the introduced distortion when projecting (ERP) the $360^{\circ}$ frames onto a 2D plane, while extracting spatial features from the videos.

\subsection{2D Video Production}
\label{subsec:2d_video}

This component takes as input the ERP frames and their associated saliency maps (extracted by the utilized saliency detection method), and produces a conventional 2D video that contains the detected salient events in the $360^{\circ}$ video. As presented in Alg. \ref{alg:2d-volumes}, this procedure starts by identifying the salient regions of each ERP frame. To form such a region in a given ERP frame, our method focuses on points of the associated saliency map that surpass an intensity value $t_1$, converts their coordinates to radians and clusters them using the DBSCAN algorithm \cite{Dbscan_1996} and a predefined distance $t_2$. Following, the salient regions that are spatially related across a sequence of frames are grouped together, thus establishing a spatial-temporally-correlated sub-volume of the $360^{\circ}$ video. For this, taking into account the entire frame sequence, our method examines whether the salient regions of the $f_{i}$ frame are close enough to the salient regions of one of the previous frames ($f_{i-1}...f_1$). If this distance is less than $t_3$, then the spatially-correlated regions over the examined sequence of frames ($f_1...f_i$) are grouped and form a sub-volume; otherwise, it creates a new sub-volume for each of the salient regions in the $f_i$ frame. Given the fact that a spatially-correlated salient region can be found in non-consecutive frames (e.g., appearing in frames $f_t$ and $f_{t+2}$), the applied grouping might result in sub-volumes that are missing one or more frames. To mitigate abrupt changes in the visual content of the formulated sub-volumes, our method adds the missing frames withing each sub-volume. Moreover, in the case that a sub-volume contains a large sequence of missing frames (higher in length than $t_4$) our method splits this sub-volume and considers that the aforementioned sequence of frames does not contain a salient event. For each sub-volume, the developed method extracts the FOV of the salient regions, thus creating a short spatio-temporally-coherent 2D video fragment. Finally, the 2D video is formed by stitching the created 2D video fragments for the different sub-volumes, in chronological order.

\algnewcommand{\LeftComment}[1]{\Statex #1 }
\begin{algorithm}[t]
\caption{2D Video Production}
\label{alg:2d-volumes}
\begin{algorithmic}[1]

\Require $N$ is the number of frames/saliency maps, $R$ the number of salient regions, $L$ the number of salient regions per frame, $S$ is the number of defined sub-volumes, $LS$ is the length (in frames) of each sub-volume, $FS$ is the number of finally formed sub-volumes, $FL$ is the length (in frames) of the finally formed sub-volumes.
\Ensure 2D Video with the salient events of the $360^{\circ}$ video

\LeftComment{\textit{``Define the salient regions in each frame by clustering salient points with intensity higher than $t_1$, using DBSCAN clustering with distance (in radians) equal to $t_2$:''}}

\For{$i=0$ to $N$}
    \State $Salient\_regions_i = F_1(Saliency\_Map_i, t_1, t_2)$
\EndFor

\LeftComment{\textit{``Define spatial-temporally-correlated 2D sub-volumes by grouping together spatially related regions (distance less that $t_3$) across a sequence of frames:''}} 

\For{$i=1$ to $N$}
    \For{$j=0$ to $L_i$}
        \State $SubVolumes_{i,j} = F_2(Salient\_regions_{i,j}, t_3)$
    \EndFor
\EndFor

\LeftComment{\textit{``Mitigate abrupt changes in the visual content of sub-volumes by adding possibly missing frames (up to $t_4$; otherwise define a new sub-volume):''}} 

\For{$m=0$ to $S$}
    \For{$n=0$ to $LS_m$}
        \State $Final\_SubVolumes_{m,n} = F_3(SubVolumes_{m,n}, t_4)$    
    \EndFor
\EndFor

\LeftComment{\textit{``Produce the 2D video by extracting the FOV for the salient regions of each finally-formed sub-volume:''}} 

\For{$k=0$ to $FS$}
    \For{$l=0$ to $FL_k$}
        \State $F_4(Final\_SubVolumes_{k,l})$
    \EndFor
\EndFor

\end{algorithmic}
\end{algorithm}

\subsection{Video Summarization}

The temporal summarization of the generated 2D is performed using a variant of the CA-SUM method \cite{10.1145/3512527.3531404}. This method integrates an attention mechanism that is able to concentrate on specific parts of the attention matrix (that correspond to different non-overlapping video fragments of fixed length) and make better estimates about their significance by incorporating knowledge about the uniqueness and diversity of the relevant frames of the video. The utilized variant of CA-SUM has been trained by taking into account also the estimates about the saliency of the video frames. In particular, the produced scores by the utilized saliency detection method (see Section \ref{subsec:saliency_prediction}) for the frames that compose the 2D video, are used to weight the extracted representations of the visual content of these frames. Hence, the utilized video summarization model is trained based on a set of representations that incorporate information about both the visual content and the saliency of each video frame. At the output, this model produces a set of frame-level importance scores. After considering the different sub-volumes of the 2D video as different video fragments, fragment-level importance scores are computed by averaging the importance scores of the frames that lie within each fragment. These fragment-level scores are then used to select the key-fragments given a target summary length $L$, by solving the Knapsack problem.

\section{Experiments}

\subsection{Datasets}

For training and evaluation of ATSal and SST-Sal, we utilized the $206$ videos of the VR-EyeTracking dataset \cite{Atsal_2020}, that were used in \cite{Xu_2018_CVPR} (two videos was excluded due to the limited clarity in its ground-truth saliency maps). VR-EyeTracking is composed of $147$ and $59$ short (up to $60$ sec.) videos that have been captured by static and moving camera, respectively. Their visual content is diverse, covering indoor scenes, outdoor activities and music shows. To train the attention model of ATSal, we also used the $85$ high definition ERP images of the Salient360! dataset \cite{Salient_2018_DATA} and the $22$ ERP images of the Sitzman dataset \cite{Sitzman_2018_Dataset}. These images include diverse visual content captured in both indoor (e.g., inside a building) and outdoor (e.g., a city square) scenes. For further evaluation, we used the $104$ videos of the Sports-360 dataset \cite{Sports360_2018_ECCV}. $84$ of them have been captured by a static camera, while the remaining ones ($18$ videos) were recorded using a moving camera. These videos last up to $60$ seconds and show activities from five different sport events (i.e., basketball, parkour, BMX biking, skateboarding and dance). Finally, for training the utilized video summarization model we used $100$ 2D videos that were produced according to the method described in Section \ref{subsec:2d_video}, and scored in terms of frame-level saliency using the methods in Section \ref{subsec:saliency_prediction}. These videos relate to $57$ videos of the VR-EyeTracking, $37$ videos of the Sports-360, and $6$ videos of the Salient360! dataset.

\subsection{Implementation Details}

The parameter $t_0$ of the decision mechanism was set equal to $0.5$. The attention model of ATSal was pre-trained for $90$ epochs using the Salient360! and Sitzman datasets. The merged dataset was composed of $2140$ ERP images, where $1840$ of them were used for training and the remaining $300$ for validation. Training was performed on mini-batches of $80$ images, using the Adam optimized with learning rate equal to $10^{-5}$ and weight decay equal to $10^{-5}$. The overall ATSal was trained using $140$ videos of the VR-EyeTracking dataset, while $66$ videos were used as validation set. Training was performed on mini-batches of $10$ frames, using the Adam optimizer with learning rate equal to $10^{-5}$ and weight decay equal to $10^{-6}$. To fine-tune SalEMA-Poles and SalEMA-Equators, we ran $20$ training epochs using mini-batches of $80$ and $10$ frames, respectively, using the Binary Cross Entropy optimizer from \cite{Salema_2019_Model} with learning rate equal to $10^{-6}$. Concerning SST-Sal, the number of hidden-layers was set equal to $9$ and the number of input-channels was set equal to $3$. Training was performed for $100$ epochs based on the Adam optimizer with a starting learning rate equal to $10^{-3}$ and an adjusting factor equal to $0.1$. $92$ videos of the VR-EyeTracking dataset were used as a training set and $55$ videos were used as a validation set. Concerning the 2D video production step, the parameters $t_1$, $t_2$, $t_3$ and $t_4$, were experimentally set equal to $150$, $1.2$, $100$ and $100$, respectively. Finally, to train the utilized variant of the CA-SUM video summarization model, deep representations were obtained for sampled frames of the videos ($2$ frames per second) using GoogleNet \cite{7298594}. The block size of the concentrated attention mechanism was set equal to $20$. The learning rate and the L2 regularization factor were equal to $5\cdot10^{-4}$, and $10^{-5}$, respectively. For network initialization we used the Xavier uniform initialization approach with gain = $\sqrt{2}$ and biases = $0.1$. Training was performed for $400$ epochs in a full-batch mode using the Adam optimizer and $80$ videos of the formed dataset; the remaining ones were used for model selection and testing. Finally, the created summary does not exceed $15\%$ of the 2D video's duration.

\subsection{Quantitative Results}
\label{subsec:quant_results}

\begin{figure}[t]
    \centering
    \includegraphics[width=\textwidth]{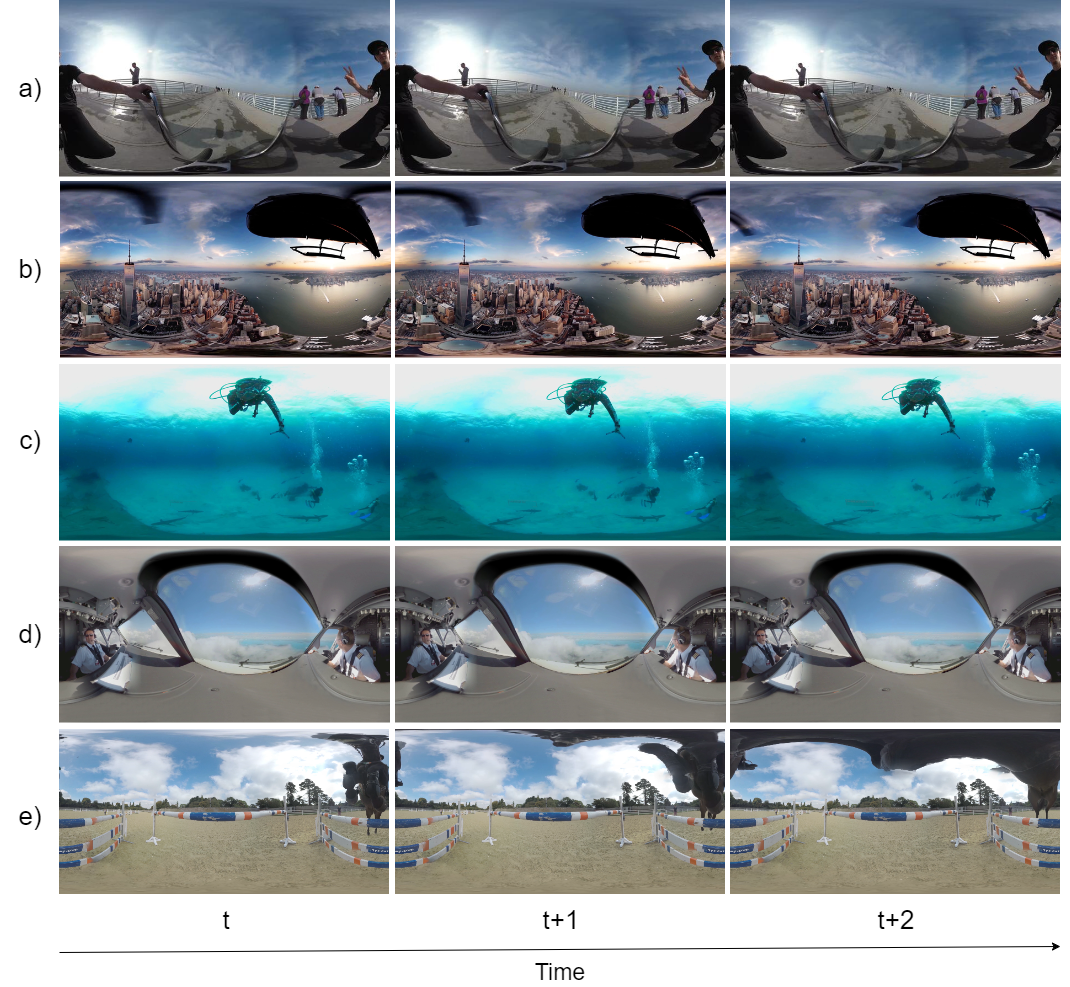}
    \caption{Examples of ERP frames from videos of the Sports-360 and VR-EyeTracking datasets, that are used by the developed decision mechanism.}
    \label{fig:decision-mech}
\end{figure}

Initially, we evaluated the accuracy of the developed decision mechanism. For this, we used $329$ videos from the VR-EyeTracking, Sports-360 and Salient360! datasets. $232$ of these videos were captured by a static camera and $97$ videos were recorder using a moving camera. The results in Table \ref{tab:decision-mech} show that our mechanism correctly classifies a video in more than $88\%$ of the cases, while its accuracy is even higher (close to $95\%$) in the case of $360^{\circ}$ videos recorded using a moving camera. The slightly lower performance in the case of static camera (approx. $86\%$) relates to mis-classifications due to changes in the visual content of the north and south regions of the $360^{\circ}$ video, caused by the appearance of a visual object right above or bellow the camera (see the example in the last row of Fig. \ref{fig:decision-mech}); however, the latter is not a desired, and thus common, case when recording a $360^{\circ}$ video using a static camera.

\begin{table}[t]
\parbox{.45\linewidth}{
\centering
\vspace{0mm}
\caption{Performance (Accuracy in percentage) of the decision mechanism.}
\label{tab:decision-mech}
\begin{tabular}{|l|c|c|c|}
\hline
                                           & \begin{tabular}[c]{@{}c@{}}Static \\ camera\end{tabular} & \begin{tabular}[c]{@{}c@{}}Moving \\ camera\end{tabular} & Total   \\ \hline
\multicolumn{1}{|c|}{Number of videos}     & 232   & 97   & 329     \\ 
\multicolumn{1}{|c|}{Correctly classified} & 200   & 92   & 292     \\ \hline
Accuracy                                   & 86.21\%   & 94.85\%   & 88.75\% \\ \hline
\end{tabular}

}
\hfill
\parbox{.45\linewidth}{
\centering
\caption{Ablation study about the use of the decision mechanism.}
\label{tab:decision-mech-atsal-sstsal}
\begin{tabular}{|l|c|c|}
\hline
                                                                                     & CC    & SIM   \\ \hline
ATSal Only                                                                           & 0.290 & 0.241 \\
SST-Sal Only                                                                         & 0.377 & 0.279 \\ \hline
\begin{tabular}[l]{@{}c@{}}Decision Mechanism \\ \& ATSal or SST-Sal\end{tabular} & \textbf{0.379} & \textbf{0.280} \\ \hline
\end{tabular}
}
\vspace{0mm}
\end{table}

\begin{table}[t]
\caption{Performance of the trained ATSal and SST-Sal models on videos of the VR-EyeTracking (upper part) and Sports-360 (lower part) datasets.}
\label{tab:saliency_detection}
\begin{center}
\begin{tabular}{@{}lcccccc@{}}
\toprule
\multicolumn{7}{c}{VR-EyeTracking}                                                                                                                                                                                                                                                                                                                                                                           \\ \midrule
\multicolumn{1}{r}{} & \multicolumn{2}{l|}{Static View Videos (55)}                                                                                  & \multicolumn{2}{l|}{Moving View Videos (11)}                                                                                  & \multicolumn{2}{l}{Total Videos (66)}                                                                                           \\
\multicolumn{1}{r}{} & CC $\uparrow$                                                           & SIM $\uparrow$                                                          & CC $\uparrow$                                                           & SIM $\uparrow$                                                          & CC $\uparrow$                                                           & SIM $\uparrow$                                                          \\
ATSal                & \cellcolor[HTML]{FFFFFF}{\color[HTML]{000000} \textbf{0.336}} & \cellcolor[HTML]{FFFFFF}{\color[HTML]{000000} \textbf{0.240}} & \cellcolor[HTML]{FFFFFF}{\color[HTML]{000000} \textbf{0.230}} & \cellcolor[HTML]{FFFFFF}{\color[HTML]{000000} \textbf{0.172}} & \cellcolor[HTML]{FFFFFF}{\color[HTML]{000000} \textbf{0.322}} & \cellcolor[HTML]{FFFFFF}{\color[HTML]{000000} \textbf{0.229}} \\
SST-Sal               & \cellcolor[HTML]{FFFFFF}0.309                                 & \cellcolor[HTML]{FFFFFF}0.167                                 & \cellcolor[HTML]{FFFFFF}0.168                                 & \cellcolor[HTML]{FFFFFF}0.106                                 & \cellcolor[HTML]{FFFFFF}0.285                                 & \cellcolor[HTML]{FFFFFF}0.157                                
                                               \\ \midrule
\multicolumn{7}{c}{Sports-360}                                                                                                                                                                                                                                                                                                                                                                               \\ \midrule
\multicolumn{1}{r}{} & \multicolumn{2}{l|}{Static View Videos (86)}                                                                                  & \multicolumn{2}{l|}{Moving View Videos (18)}                                                                                  & \multicolumn{2}{l}{Total Videos (104)}                                                                                          \\
\multicolumn{1}{r}{} & CC $\uparrow$                                                           & SIM $\uparrow$                                                          & CC $\uparrow$                                                           & SIM $\uparrow$                                                         & CC $\uparrow$                                                           & SIM $\uparrow$                                                          \\
ATSal                & \cellcolor[HTML]{FFFFFF}0.270                                 & \cellcolor[HTML]{FFFFFF}0.251                                 & \cellcolor[HTML]{FFFFFF}0.270                                 & \cellcolor[HTML]{FFFFFF}0.243                                 & \cellcolor[HTML]{FFFFFF}0.270                                 & \cellcolor[HTML]{FFFFFF}0.249                                 \\
SST-Sal               & \cellcolor[HTML]{FFFFFF}\textbf{0.464}                        & \cellcolor[HTML]{FFFFFF}\textbf{0.372}                        & \cellcolor[HTML]{FFFFFF}\textbf{0.273}                        & \cellcolor[HTML]{FFFFFF}\textbf{0.283}                        & \cellcolor[HTML]{FFFFFF}\textbf{0.436}                        & \cellcolor[HTML]{FFFFFF}\textbf{0.358}                        \\
                     & \multicolumn{1}{l}{}                                          & \multicolumn{1}{l}{}                                          & \multicolumn{1}{l}{}                                          & \multicolumn{1}{l}{}                                          & \multicolumn{1}{l}{}                                          & \multicolumn{1}{l}{}     

\end{tabular}
\end{center}
\vspace{-6mm}
\end{table}

Following, we assessed the performance of the ATSal and SST-Sal methods for saliency detection. The evaluation was based on the Pearson Correlation Coefficient (CC) and the Similarity (SIM) measures, as proposed in \cite{Metrics_2019}. CC calculates the linear correlation between the distributions in a pair of automatically extracted and ground-truth saliency maps. SIM quantifies the similarity between the aforementioned distributions after seeing them as histograms. The results of the conducted experiments are reported in Table \ref{tab:saliency_detection}. As can be seen, ATSal performs better on videos of the VR-EyeTracking dataset and SST-Sal is more effective on videos of the Sports-360 dataset; so, practically we have a tie between these two methods. However, the performance of ATSal on videos captured by a moving camera (denoted as ``moving view videos'' in Table \ref{tab:saliency_detection}) is significantly higher than the performance of SST-Sal on the VR-EyeTracking dataset, and slightly lower but comparable with the performance of SST-Sal on the Sports-360 dataset. The exact opposite can be observed in the case of videos recorded using a static camera (denoted as ``static view videos'' in Table \ref{tab:saliency_detection}). The SST-Sal method performs clearly better than ATSal on the Sports-360 dataset and comparatively good on the VR-EyeTracking dataset. Based on these findings, we integrate both ATSal and SST-Sal in the saliency detection component of the system; the former is used to analyze $360^{\circ}$ videos captured by a moving camera and the later processes $360^{\circ}$ videos recorded using a static camera. 

Finally, to evaluate the impact of the utilized decision mechanism, we formed a large set of test videos (by merging the $104$ test videos of the Sports-360 dataset with the $66$ test videos from the VR-EyeTracking dataset) and considered three different processing options: a) the use of ATSal only, b) the use of SST-Sal only, and c) the use of both methods in combination with the developed decision mechanism. The results in Table \ref{tab:decision-mech-atsal-sstsal} document the positive impact of the decision mechanism, as its use in combination with the integrated saliency detection methods results in higher performance.

\begin{figure}[t]
    \centering
    \includegraphics[width=\textwidth]{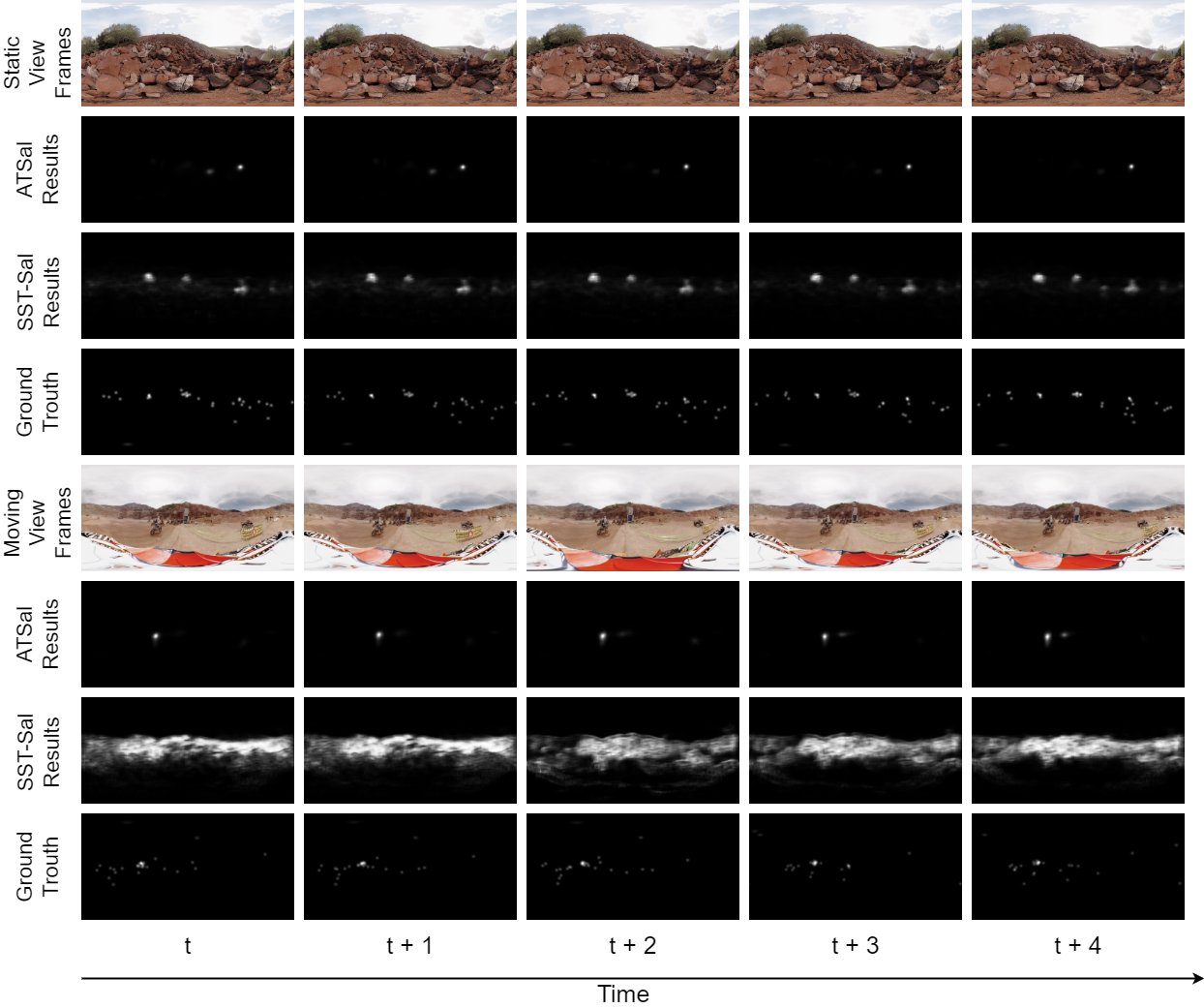}
    \caption{Qualitative comparisons between the output of the ATSal and SST-Sal methods on frame sequences of videos from the Vr-EyeTracking dataset.}
    \label{fig:saliency}
\end{figure}

\subsection{Qualitative Results}

Figure \ref{fig:decision-mech} shows examples of ERP frames that are processed by the decision mechanism. In the first three cases (rows a), b), and c) of Fig. \ref{fig:decision-mech}), the mechanism detects changes in the north and south regions of these frames, and correctly identifies the use of a moving camera for video recording. However, for the images in d) and e) rows of Fig. \ref{fig:decision-mech} our mechanism fails to make a correct decision. In the former case (row d), despite the fact that the video was recorded from the cockpit of a moving helicopter, the mechanism was not able to detect sufficient motion in the regions that focuses on. In the latter case (row e), the appearance of a horse right above the static camera led to noticeable changes in the observed regions, thus resulting in the erroneous detection of a moving camera.

Figure \ref{fig:saliency} presents two sequences of ERP frames, the produced saliency maps by ATSal and SST-Sal, and the ground-truth salience maps. Starting from the top, the first frame sequence was extracted from a $360^{\circ}$ video captured using a static camera. From the associated saliency maps we can observe that SST-Sal performs clearly better compared to ATSal and creates saliency maps that are very close to the ground-truth; on the contrary, ATSal fails to detect several salient points. For the second frame sequence, which was obtained from a $360^{\circ}$ video recorded using a moving camera, we see the exact opposite behavior. ATSal defines saliency maps that are very similar with the ground-truth, while the saliency maps of SST-Sal method contain too much noise. The findings of our qualitative analysis are aligned with our observations in the quantitative analysis, justifying once again the use of ATSal (SST-Sal) as the best option for analysing $360^{\circ}$ videos captured using a moving (static) camera.

\begin{figure}[t]
    \centering
    \includegraphics[width=\textwidth]{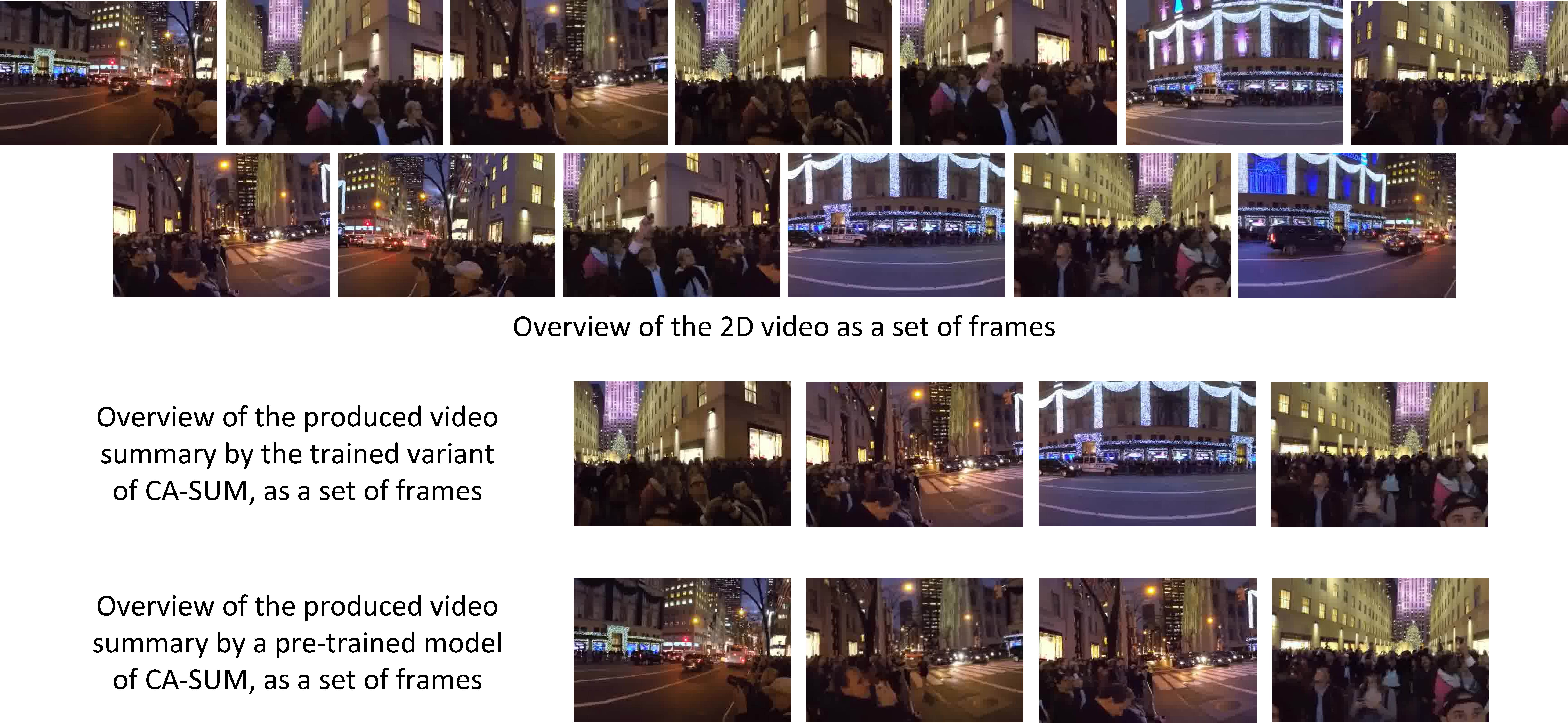}
    \caption{Qualitative comparisons between the output of the ATSal and SST-Sal methods on frame sequences of videos from the VR-EyeTracking dataset.}
    \label{fig:summarization_examples}
\end{figure}

Figure \ref{fig:summarization_examples} gives a frame-based overview of the produced 2D video for a $360^{\circ}$ video after selecting one frame per shot (shots are directly related to the defined sub-volumes of the $360^{\circ}$ video, as described in Section \ref{subsec:2d_video}), and presents the produced summaries by two video summarization methods. The summary at the top was created by the trained saliency-aware variant of the CA-SUM method using videos of the VR-EyeTracking and Sports-360 and Salient360! datasets. The summary at the bottom was created by a pre-trained model of CA-SUM using conventional videos from the TVSum dataset \cite{7299154} for video summarization. As can been seen, the trained variant produces a more complete and representative video summary after including parts of the video showing the gathered people in a square with a Christmas tree, the persons right in front of the avenue that take some photos, and the illuminated shopping mall behind the avenue. On the contrary, the pre-trained CA-SUM model focuses more on fragments of the video showing the avenue and ignores video parts presenting the shopping mall. This example shows that taking into account the saliency of the visual content is important when summarizing $360^{\circ}$ videos, as it allows the production of more representative and thus useful video summaries.

\section{Conclusions}
In this paper, we described an integrated solution for summarizing $360^{\circ}$ videos. To create a video summary, our system initially processes the $360^{\circ}$ video in order to indicate the use of a static or moving camera for its recording. Based on the output, the $360^{\circ}$ video is then forwarded to one of the integrated state-of-the-art methods for saliency detection (ATSal and SST-Sal), which produce a set of saliency maps. The latter are utilized to form spatiotemporally-correlated sub-volumes of the $360^{\circ}$ video that relate to different salient events, and build a conventional 2D video that shows these events. Finally, a saliency-aware variant of a state-of-the-art video summarization method (CA-SUM) analyses the produced 2D video and formulates the video summary. Quantitative and qualitative evaluations using two datasets for saliency detection in $360^{\circ}$ videos (VR-EyeTracking, Sports-360), demonstrated the performance of different components of the system and documented their relative contribution in the $360^{\circ}$ video summarization process.
%\bibliographystyle{splncs04}
%\bibliography{bibliography}
%

\end{document}